# A Clustering Approach to Solving Large Stochastic Matching Problems

Milos Hauskrecht
Department of Computer Science
Mineral Industries Building
University of Pittsburgh
Pittsburgh, PA 15260
*milos@cs.pitt.edu*

Eli Upfal
Department of Computer Science, Box 1910
Brown University
Providence, RI 02912
*eli@cs.brown.edu*

## Abstract

In this work we focus on efficient heuristics for solving a class of stochastic planning problems that arise in a variety of business, investment, and industrial applications. The problem is best described in terms of future *buy* and *sell* contracts. By buying less reliable, but less expensive, buy (supply) contracts, a company or a trader can cover a position of more reliable and more expensive sell contracts. The goal is to maximize the expected net gain (profit) by constructing a close to optimum portfolio out of the available buy and sell contracts. This stochastic planning problem can be formulated as a two-stage stochastic linear programming problem with recourse. However, this formalization leads to solutions that are exponential in the number of possible failure combinations. Thus, this approach is not feasible for large scale problems. In this work we investigate heuristic approximation techniques alleviating the efficiency problem. We primarily focus on the clustering approach and devise heuristics for finding clusterings leading to good approximations. We illustrate the quality and feasibility of the approach through experimental data.

## 1 Introduction

While many practical decision and planning problems can be modeled and solved as deterministic optimization problems, a significant portion of real world problems is further complicated by the presence of uncertainty in the problem parameters. In solving such problems one not only faces the complexity of the original optimization problem but also the complexity arising from random fluctuations of parameters and global criteria summarizing and quantifying all possible random behaviors. The challenge here is to devise methods capable of efficiently solving large scale instances of such problems.

In this paper we investigate a class of stochastic planning problems that arise in many business, investment and industrial applications. We term this problem *stochastic contract matching* and formulate it in terms of optimizing a portfolio of future (call) contracts [15] (the same optimization problem comes up in a variety of other applications such as insurance contracts). A *call* contract is an option to buy a given commodity at a given price. A call contract has a default clause specifying the penalty that the seller of the contract pays if the contract cannot be satisfied. In volatile markets of commodities such as energy (gas, electricity) and communication bandwidth there is a big price spread between "reliable" contracts with high default penalties and "less reliable" contracts with relatively negligible penalties. By buying a collection of less reliable, but less expensive contracts a trader can cover, at a significant profit, a position of expensive, reliable, contracts that he had sold to clients.

In our formalization a *buy* contract is a call contract bought by the trader (typically a less expensive and less reliable contract), and a *sell* contract is a call contract sold by the trader to a client (typically, a more expensive and more reliable contract). Each buy contract can cover one of a set of sell contracts. The goal is to maximize the expected net gain (profit) by constructing a close to optimum portfolio out of the available buy and sell contracts. The gain of the portfolio is the revenue from selling the "sell" contract minus the cost of purchasing the "buy" contracts and the penalties for uncovered sell contracts [8]. (In practice, the penalty on "reliable" contracts is so high that a trader must satisfy all of them, possibly through an expensive "spot" market).

**Example:** Consider a problem of trading communication bandwidth through unreliable satellite and/or ground transmitter equipment and its channels. Our goal is to find the best combination of lease (buy) and sell contracts maximizing the expected value (profit), by taking into an account a probability of equipment failures, flexibility of equipment coverage, profits/costs for selling/buying respective contracts and penalties for breaching sell contracts.

The stochastic contract matching problem represents a stochastic planning problem with two decision steps: (1) an allocation problem, deciding which contracts to buy and sell, and (2) a matching problem, where the decision about the best coverage of sell contracts after observing the actual failure configuration is made. The problem can be formulated as a two stage stochastic programming problem with recourse [4, 1]. While there are efficient techniques to solve the deterministic version of the matching problem [13, 1], the stochastic version becomes exponential in the number of randomly fluctuating elements. It is this aspect we address in our work.

There has been extensive research in AI in recent years on solving stochastic planning problems with large action and state spaces and variety of techniques for reducing the complexity (typically exponential in the number of components) of these problems have been proposed [3, 7, 6, 5, 9, 12, 2, 11]. However, all these works assume a fixed structure and a fixed parameterization of the planning problem. The unique aspect of our planning problem is that the underlying topology characterizing the problem can vary and it is itself subject to random changes and fluctuations (due to failures). The optimal decisions must account for these effects.

We focus on and propose efficient heuristic approximation techniques to solve the stochastic matching problem. In particular, we develop a novel clustering approach. The idea of the clustering technique is to reduce the number of stochastic configurations to be considered in the optimization by aggregating similar configurations. Ideally we would like to get the smallest possible number of clusters leading to the best approximation. Computing the optimal clustering is as hard as solving the original problem. The clustering approach and associated heuristics we propose are computationally feasible and lead to lower (upper) bound approximations of the optimal solution. The quality and computational efficiency of the approach are demonstrated empirically on experimental data.

## 2 The Model

We have two sets of contracts for commodities (products, services etc.):

- Buy contracts — a right to one unit of a service or commodity. The contract has a price $R^b$, and a known failure probability.

- Sell contracts — an obligation to deliver one unit of a service or commodity. The contract has a price $R^s$, and a penalty $\bar{R}^s$ for not satisfying the contract.

In addition, we have a function defining which buy contracts can satisfy a sell contract (Figure 1). Note that since each buy contract can cover only one sell contract, exercising the contracts is equivalent to a matching of buy and sell contracts.

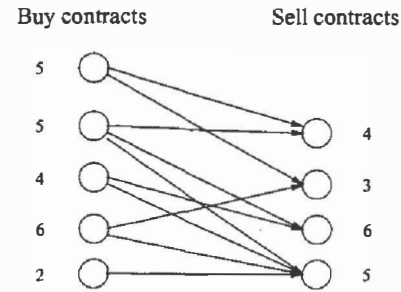

Figure 1: An example of contract asset matching. Nodes corresond to different types of contracts. Links represent one-to-one matchings between buy and sell assets covered by contracts. Numbers reflect the limits for each type of contract.

The objective is to find the optimal position (portfolio) of contracts (available on the market) while optimizing an objective function that takes into an account the possibility of various failures, and subsequent contract breaches. To incorporate uncertainty and possible failures we focus on the *expected value measure*, where we want to find a contract position leading to the optimal expected profits. Other more complex measures, e.g. incorporating different risk preferences of an investor, are possible as well.

## 3 Formulating the optimization problem

### 3.1 Notation

Let $q$ be a number of different buy contracts (with possibly different price or buy asset links) and $n_u$ be the number of contracts of type $u$. Let $k$ be a number of sell contract types and $m_i$ the number of such contracts of type $i$. Both buy and sell contracts have price; $R_u^b$ denotes the price of a buy contract of type $u$, $R_i^s$ the price of sell contract of type $i$. $\bar{R}_i^s$ the penalty we have to pay for not satisfying a sell contract $i$. (The penalty for not satisfying a buy contract is 0.) The number of contracts of each type we can hold is restricted and satisfies: $0 \leq n_u \leq C_u^b$ for all $u = 1, 2, \cdots, q$ and $0 \leq m_i \leq C_i^s$ for all $i = 1, 2, \cdots, k$, where $C$ denotes upper limits on the number of buy and sell contracts. Enforcing 0 lower bound limits ensures that no short-selling of contracts can occur.

### 3.2 Deterministic matching problem

The decision problem consists of two choices. First we select a combination of buy and sell contracts. Second, after observing the actual failure configuration, we decide how to match different buy and sell contracts. Here we focus on the second step and its optimal solution.

Let $S = s_1 s_2 \cdots s_u \cdots s_q$ be an observed failure combination, such that $s_u = 1$ if buy contracts $u$ did not fail and $s_u = 0$ if they failed. Suppose penalties for breaching the contract are expressed in terms of negative rewards, the matching problem can be formulated as a linear optimization problem:

$$Q(\mathbf{n}, \mathbf{m}, S) = \max_{\mathbf{j}} \left[ -\sum_{i=1}^{k} \sum_{u=1}^{q} j_u^i \bar{R}_i^s \right],$$

subject to constrains:

$$m_i - \sum_{u=1}^{q} j_u^i \geq 0 \text{ for all } i,$$

$$s_u n_u - \sum_{i=1}^{k} j_u^i \geq 0 \text{ for all } u,$$

$$j_u^i \geq 0 \text{ for all } u, i.$$

In this LP, $n_u$ represents the number of buy assets of type $u$ ($\mathbf{n} = \{n_1, n_2, n_u, \cdots n_q\}$), $m_i$ the number of sell contracts of type $i$ ($\mathbf{m} = \{m_1, m_2, \cdots, m_i, \cdots m_k\}$), and $\mathbf{j}$ is a vector (set) of variables $j_u^i$ representing the number of units of asset to be distributed from $u$ to $i$ along connection $(u, i)$ if it exists (the variable $j_u^i$ can be omitted if the connection is not present). Values of $\mathbf{n}$, and $\mathbf{m}$ are fixed and constants and not subject to optimization. The objective function $Q$ essentially attempts to minimize losses by covering sell contracts with highest penalties. There are two sets of constraints. The first set assures that the number of buy assets actually matched does not exceed the number of sell assets (covered by sell contracts). The second set of constraints assures that we distribute only assets available on the buy side.

The above matching problem (with fixed supplies and demands) is a special case of a Hitchcock problem [13]. An interesting property of the problem is that for integral constraints, its basic feasible solutions are integral.[1]

### 3.3 Contract portfolio optimization

Our ultimate goal is to find the combination of $n_u$ and $m_i$ values leading to the best (maximum) expected profits. The problem can be formulated as a two stage linear program with recourse (see e.g. [1]):

$$V = \max_{\mathbf{n,m}} \left\{ E_S(Q(\mathbf{n}, \mathbf{m}, S)) - \sum_{u=1}^{q} n_u R_u^b + \sum_{i=1}^{k} m_i (R_i^s + \bar{R}_i^s) \right\}$$

subject to

$$C_u^b \geq n_u \geq 0 \text{ for all } u;$$
$$C_i^s \geq m_i \geq 0 \text{ for all } i,$$

where $E_S(Q(\mathbf{n}, \mathbf{m}, S))$ is the expectation of $Q$ for different failure combinations. The two-stage problem can be expanded into a linear program of the form:

$$V = \max_{\mathbf{n,m,j}} \left\{ -\sum_{u=1}^{q} n_u R_u^b + \sum_{i=1}^{k} m_i (R_i^s + \bar{R}_i^s) \right.$$
$$\left. - \sum_{v=1}^{r} \sum_{i=1}^{k} \sum_{u=1}^{q} j_{u,v}^i [p(S^v) \bar{R}_i^s] \right\}$$

subject to constraints:

$$m_i - \sum_{u=1}^{q} j_{u,v}^i \geq 0 \text{ for all } i, v;$$

$$s_u^v n_u - \sum_{i=1}^{k} j_{u,v}^i \geq 0 \text{ for all } u, v;$$

$$j_{u,v}^i \geq 0 \text{ for all } u, v, i;$$

$$C_u^b \geq n_u \geq 0 \text{ for all } u;$$

$$C_i^s \geq m_i \geq 0 \text{ for all } i,$$

where $v$ ranges over all possible combinations of failures $v = 1, 2, \cdots, r$; $p(S^v)$ is a probability of a failure combination $v$, and all variables in the deterministic matching subproblem are also indexed by $v$.

A similar LP can be constructed to evaluate a specific buy and sell contract position under the assumption of the optimal matching. The difference between the evaluation and optimization is that $\mathbf{n}, \mathbf{m}$ are either variables (optimization) or fixed values (evaluation). Note, hovewer, that in terms of the number of failure combinations the evaluation task is comparable to the optimization.

The two-stage problem with a recourse (or its expanded version) offers a special structure allowing more specific optimization techniques to be applied to solve it. The methods include basic (1-cut) or multicut L-shaped methods [14] and inner linearization methods. For a survey of applicable techniques see [1]. While basic feasible solutions of the deterministic matching LP with integral constraints are integral, the integrality of a two-stage solution remains an interesting open issue.[2]

The apparent drawback of solving the contract optimization problem is the curse of dimensionality; the complexity of the LP formulation is in the worst case exponential in the number of components that can fail and $r = 2^q$. Note that the curse of dimensionality affects also the evaluation task in which we want to compute the expected value of a fixed set of buy and sell allocations under the optimal after-failure matching. Thus it is hard to even evaluate a fixed allocation. One possibility to alleviate this problem is to assure (via various structural restrictions) that the number of failure combinations is small and polynomial. Then the exact solution can be obtained efficiently. Another possibility is to apply various heuristics leading to efficient solutions.

---

[1]The existence of an integral solution is a consequence of total unimodularity property of the matrix defining an LP[13].

[2]Our experiments with two-stage problems always lead to integral solutions. However, we currently do not have a theoretical proof of the property, or a counterexample.

## 4 Greedy approaches

One way to solve the contract optimization problem is to apply greedy heuristics in which the solution is constructed incrementally such that partial matchings with highest expectations are preferred and selected first.

#### 4.0.1 Pairwise greedy

There are various versions of the greedy algorithm. The simplest algorithm checks expected profits for all possible buy-sell matchings, orders the matchings and builds the solution incrementally by selecting contracts corresponding to the best remaining buy-sell pair (according to the ordering). The expected profit for matching a pair of contracts $(u, i)$ is:

$$V(u, i) = -R_u^b + (1 - p_u)R_i^s + p_u \bar{R}_i^s$$

where $p_u$ is the failure probability of a buy contract $u$. During the solution-building process, the number of buy and sell contracts should never exceed capacity constraints. The process stops when there are no additional pairs satisfying capacity constraints or when expected profits of remaining pairs are negative.

The drawback of the above algorithm is that it does not allow diversification. In other words, the algorithm never recommends buying two or more buy contracts to cover one sell contract and this despite the fact that this choice can increase the overall value of the solution.

#### 4.0.2 Diversified greedy

A partial remedy to the above problem is to diversify individual sell contracts across different buys. From the viewpoint of a sell contract $i$ only, we want to select a subset $B_z^i$ of all buy assets incident on $i$ (denoted $B^i$), leading to the best value:

$$V(B_z^i, i) = -\left[\sum_{u \in B_z^i} R_u^b\right] + (1 - p_{B_z^i})R_i^s + p_{B_z^i}\bar{R}_i^s,$$

where $p_{B^i}$ is the probability of all buy contracts in $B_z^i$ failing. The best subset can be found either through an exhaustive search or by setting up a linear program similar to the original linear program. In both cases the solution is exponential in $|B_i|$. Thus assuming that $\max_i |B^i|$ is small the local diversification can be performed exactly.

| method<br>contracts | pairwise<br>greedy | diversified<br>greedy | exact |
|---|---|---|---|
| buy (n) | (5 5 0 0 0) | (5 5 2 5 2) | (5 5 4 5 2) |
| sell (m) | (2 3 5 0) | (2 3 5 2) | (4 3 4 5) |

Table 1: Comparision of buy and sell allocations for two greedy methods and the optimal solution.

Different buy contracts can be used to cover one or more sell contracts. To resolve possible conflicts among different sell contracts (buy assets are shared) we select greedily the sell contract (and its best buy combination) with the highest expected value and allocate the maximum available capacity to it. We repeat the allocation process while dynamically adjusting capacity constraints and stop when capacity constraints are saturated or when none of the best combinations comes with a positive expected value.

The new greedy method decreases a chance of not satisfying a sell contract by using a multiple buy coverage, thus improving on the pairwise greedy method. Unfortunately, it also ignores the possibility of using one buy contract to diversify simultaneously more sell contracts which is one the key features of our problem. Table 1 illustrates the differences among the two greedy methods and the optimal solution on the problem from Figure 1 with 5 different types of buy contracts and 4 types of sell contracts. The diversified greedy method chooses multiple different buy contracts as compared to the pairwise greedy allowing to cover one sell contract with multiple buys. In addition, more sell contracts are sold since positive gains can appear as a result of diversification and multiple coverage. However, in the optimal solution a buy contract can be also used to diversify many sell contracts, leading to the increase in the number of sell contracts.

## 5 Approximation based on clustering

To improve on the two greedy methods we develop an alternative approach – cluster-based approximation. The idea of our cluster-based approximation is to: (1) restrict the number of failure configurations $r$ considered in the optimization problem (LP) and (2) approximate the effect of all other failure combinations only through configurations in the restricted set. The probability of each configuration in the restricted set is modified accordingly and covers all configurations it replaced. A set of failure combinations substituted by the same representative configuration is called a *cluster*; the configuration representing a cluster is a *cluster seed*.

The actual profits for a specific contract position depend on the number of failures that occurred. In general, more failures reduce our ability to satisfy sell contracts and thus tend to decrease the profits when compared to the situation with less failures. By disregarding some of the failure combinations and substituting them with combinations with more failures one obtains a lower bound estimate of the expected value of a given portfolio of contracts. Thus, clustering failures such that combinations are only replaced with combinations with more failures leads to a lower bound approximation. Small (polynomial) number of such clusters considered in evaluation (optimization) then leads to a polynomial lower bound solution. Analogously, by substituting failure combination with configurations with smaller number of failures one obtains an efficient upper bound solution estimate. This is the key idea of our approach.

To fully develop the clustering idea we need to:

1. define a clustering method that for a given set of seed failure combinations leads to a lower (upper) bound estimate of the optimal expected value;

2. compute a probability distribution of these clusters;

3. choose (build) a combination of cluster seeds defining the approximation.

### 5.1 Upper and lower bound clustering

Let $S = s_1 s_2 \cdots s_q$ denotes a specific failure combination, such that $s_u = 0$ if buy contracts $u$ failed and $s_u = 1$ otherwise.

**Definition 1** *Let $S_1$ and $S_2$ be two failure combinations. We say that $S_1$ failure-dominates $S_2$ if $s_u^1 = 0$ whenever $s_u^2 = 0$ holds. We say that $S_1$ non-failure-dominates $S_2$ when $s_u^1 = 1$ holds whenever $s_u^2 = 1$.*

It is easy to see that failure and non-failure dominance are closely related: a configuration $A$ failure-dominates $B$, iff $B$ non-failure dominates $A$.

To guarantee a lower bound estimate of the expected value we substitute a specific failure combination only with a failure combination that failure-dominates it. Analogously, to obtain an upper bound estimate a failure combination can be substituted only by a failure combination that non-failure-dominates it. Other substitutions may violate the bounds. To assure the whole configuration space is always covered, our cluster set always includes all-fail and all-no-fail combinations.

In essence, a clustering partitions the space of failure configurations. The number of possible partitionings is exponential. In this work, we develop a special form of clusterings that are defined in terms of the seed set orderings. The advantage of the clustering is that it reflects the symmetry of failure and non-failure dominance and it can be used to obtain both bounds.

**Definition 2** *A clustering is defined by a fixed ordering of seed configurations $W = \{S_1, S_2, \cdots S_r\}$, such that $S_1$ is the all-no-fail combination, $S_r$ is the all-fail combination, and for all pairs $S_i, S_j$ s.t. $i < j$, holds that $S_i$ does not failure dominate $S_j$. In the **lower bound clustering**, a configuration belongs to the first cluster (seed) that failure-dominates it, starting from $S_1$. In the **upper bound clustering**, a configuration belongs to the first cluster that non-failure dominates it, starting from $S_r$ and checking seeds in $W$ in the reverse order.*

### 5.2 Computing probabilities of clusters

Once the clustering is known, the next step is to compute the probability mass of each cluster. Here, we assume the lower bound clustering, the upper bound is a dual problem.

Let $W = S_1, S_2, \cdots S_r$ be an ordered set of seeds defining the clustering and let $\cap_f$ denotes a *failure overlap operator*, $S' = S_i \cap_f S_k$, such that for all $u = 1, \cdots q$ holds:

$$s'_u = \begin{cases} 0 & \text{if } s_u^i = s_u^k = 0 \\ 1 & \text{otherwise} \end{cases}$$

Then the probability of a cluster $\text{cl}(S_j)$ is by the inclusion-exclusion sum:

$$\begin{aligned} p(\text{cl}(S_j)) &= p(\text{fds}(S_j)) - \sum_{i=1}^{j-1} p(\text{fds}(S_j \cap_f S_i)) \quad (1) \\ &+ \sum_{i=1}^{j-2} \sum_{k=i+1}^{j-1} p(\text{fds}(S_j \cap_f S_i \cap_f S_k)) - \cdots \end{aligned}$$

$\text{fds}(S)$ is a set of all configurations failure-dominated by a configuration $S$ and $p(\text{fds}(S))$ its probability mass. $p(\text{fds}(S))$ equals the marginal probability of all non-failed buy contracts in the configuration. For independent failures it equals:

$$p(\text{fds}(S)) = \left[ \prod_{u_n} (1 - p_{u_n}) \right],$$

where $u_n$ ranges over all buy contracts that did not failed in $S$.

To obtain the probability of a cluster $\text{cl}(S_j)$ for $W$ (equation 1) we modify $p(\text{fds}(S_j))$ by substracting the probability mass already captured by other cluster seeds $S_1, S_2, \cdots, S_{j-1}$. This assures that the probability of any failure combination is not counted twice.

#### 5.2.1 Approximations of cluster probabilities

The equation 1 gives us a recipe to compute the probability distribution of a given set of clusters consistent with a lower bound approximation. However, in order to obtain efficient approximation this computation must be efficient. Assuming all marginal probabilities are efficiently computable, the inclusion-exclusion (IE) which requires to evaluate all possible configuration overlaps represents the main difficulty.

To resolve this problem we compute upper and lower bound estimates of cluster probabilities using standard approximations of the IE problem. The solution is to consider only a limited number of intersections, such that we end with a negative sign correction to assure a lower bound and a positive sign to obtain an upper bound. Let $\widehat{p}(\text{cl}(S_j)) \leq p(\text{cl}(S_j))$ be a lower bound probability of a cluster $\text{cl}(S_j)$ (for $W$), obtained via IE approximation. As every cluster includes at least its seed, its probability mass can be lower bounded by:

$$p'(\text{cl}(S_j)) = \max[p(S_j); \widehat{p}(\text{cl}(S_j))],$$

where $p(S_j)$ is the joint probability of a configuration $S_j$. To assure that probabilities of all clusters sum to one, we add all unaccounted probability mass (can appear due to the approximation of the IE problem) to the cluster seeded by all-fail combination ($S_r$ configuration). That is:

$$p'(S_r) = 1 - \sum_{j=1}^{r-1} p'(\text{cl}(S_j)).$$

An alternative to inclusion-exclusion approach is to estimate cluster probabilities directly using Monte-Carlo techniques. Note that this approach can be more convenient also in the case when marginal probabilities needed for IE approximations are hard to compute.

### 5.3 Finding good clusterings

The last challenge is to devise techniques for finding a clustering leading to a good approximation of the optimal solution. This problem consists of two closely related subproblems: (1) finding the best clustering (the best order) of a fixed set of seed points, and (2) choosing the set of seed points defining the approximation. In general, it is hard to solve any of these from scratch in one-shot. Thus instead, we focus on incremental methods improving clusterings gradually, while exploiting the previously built approximation.

### 5.4 Best seed ordering

Let $A$ be a set of seed points. A clustering is defined by an ordering $W$ of seed points in $A$ (definition 2) and divides (clusters) the space of all failure configurations. As pointed out earlier, there can be different orderings of elements in $A$ and in the worst case the number is exponential. In general, the solutions (allocations of $\mathbf{m}, \mathbf{n}$) corresponding to different orderings may be different. However, despite this fact, it is very often possible to improve the ordering of seeds by examining Q-values of a two-stage linear program. This idea is captured in the following theorem.

**Theorem 1** *Let $S_i$ and $S_j$ be two seeds in $W$ such that $i < j$. Let $V^W$ be the optimal value for $W$. If $Q^W(\mathbf{m^W}, \mathbf{n^W}, S_i) < Q^W(\mathbf{m^W}, \mathbf{n^W}, S_j)$, then there is an ordering $W'$ such that $S_j$ preceeds $S_i$ in this ordering and $V^{W'} \geq V^W$.*

**Proof** Let $S_{ij} = S_i \cap_f S_j$ be a failure overlap of $S_i$ and $S_j$. For the ordering $W$ the probability mass of $S_{ij}$ can belong (may be in part) to $S_i$; it never belongs to $S_j$. Given the fact that $\mathbf{m^W}, \mathbf{n^W}$ are the optimal allocations for $W$, such that $Q^W(\mathbf{m^W}, \mathbf{n^W}, S_i) < Q^W(\mathbf{m^W}, \mathbf{n^W}, S_j)$, assigning the probability mass of the overlap to $S_j$ (for the same allocation $\mathbf{m^W}, \mathbf{n^W}$) must lead to a better expected value. Note that the condition $Q^W(\mathbf{m^W}, \mathbf{n^W}, S_i) < Q^W(\mathbf{m^W}, \mathbf{n^W}, S_j)$ implies that $S_j$ does not failure-dominate $S_i$ (if it would, the $Q$ value of $S_j$ cannot be larger), thus the new ordering exists and is valid. □

The theorem gives rise to a simple but very effective iterative improvement procedure for a set of seed points $A$: select an initial seed ordering, solve the approximation problem, compute $Q$ values for every seed, sort them according to $Q$-values, and solve the problem repeatedly until no changes in the seed order are observed. Note that sorting seed points according to their Q-values works also for the upper bound case. Intuitively, in the upper bound case we want to find the clustering that leads to the smallest (tightest) upper bound. As upper bounds use the reverse ordering of $W$, sorting the seeds according to their Q-values guarantees to improve also the upper bound.

Although the above iterative procedure may not lead to the globally optimal clustering it always guarantees an improvement and is easy to implement. To search for the globally optimal solution, combinatorial optimization techniques such as Metropolis algorithm [10] allowing to scan a space of seed orderings can be combined with the heuristic.

### 5.5 Selecting cluster seeds

Our ultimate goal is to approximate the optimal solution. As it is hard to guess a good set of cluster seeds in one step, we focus on the incremental approach in which we improve the approximation by gradually refining the cluster set.

Intuitively, both cluster probabilities and Q-values of cluster seeds influence the expectation and thus heuristics should reflect both. To capture the effect of probabilities we use the following heuristic: to add a new seed, we first choose the cluster with the largest probabilistic mass not accounted for by the seed configuration itself, and after that we choose a configuration from within the cluster randomly (according to the probability distribution). Let $p(\text{cl}(S_i))$ be a probability of a cluster defined by a seed $S_i$ or its estimate and $p(S_i)$ the probability of a seed configuration itself. Then we define the value of a cluster $\text{cl}(S_i)$ as: $H(\text{cl}(S_i)) = p(\text{cl}(S_i)) - p(S_i)$. The heuristic selects the cluster with the highest value of $H$, thus splitting the cluster with the largest potential to improve the approximation.

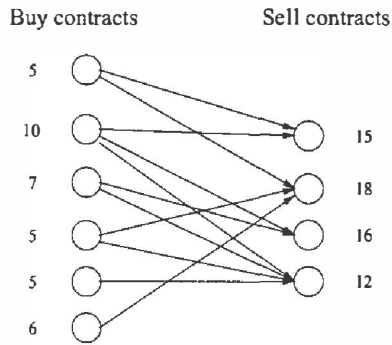

Figure 2: Admissible matching for the problem with 6 buy and 4 sell contract types used in experiments. Numbers indicate capacity limits for each contract type.

To incorporate the effect of Q-values we apply the reordering heuristics (seeds are sorted according to the Q-values for the last seeds set) after every step. The objective of is to improve the clustering by considering a newly added seed and its $Q$-value.

### 5.6 Experiments

We have tested the incremental strategy together with the two heuristic refinements on a problem with 6 buy sites and 4 sell sites. Figure 2 shows all admissible matchings between buy and sell contracts. Figure 3 plots values of lower bound approximations obtained by gradually increasing the number of clusters. Averages of 10 trials are shown for each combination of methods. As the values represent lower bounds, a higher value indicates better approximation. For comparison, we also plot expected values for the optimal allocation and allocations for the diversified and pairwise greedy methods. The best performance was obtained by the combination of the two heuristics - probability based seed selection and reclustering (reordering of seeds) based on Q-values. On the other hand, the worst performing method selects new seed configurations uniformly at random, with no reclustering. The other two choices, came in between, with probability-based heuristics edging the reclustering.

Figure 4 shows the average running times of approximations for different number of clusters. The only significant difference between the methods we observed is due to reclustering heuristics which reevaluates the cluster seed order and improves the seed ordering locally (for each cluster size). The two curves shown average the running times of methods with and without reclustering (reordering of seeds). To solve a two stage LP problem we use the VNI linear programming package. In contrast to cluster approximations the optimal solution was obtained in 352 minutes. Thus, using the combinations of our heuristics we were able to obtain approximations very close to the optimal value in a significantly shorter time.

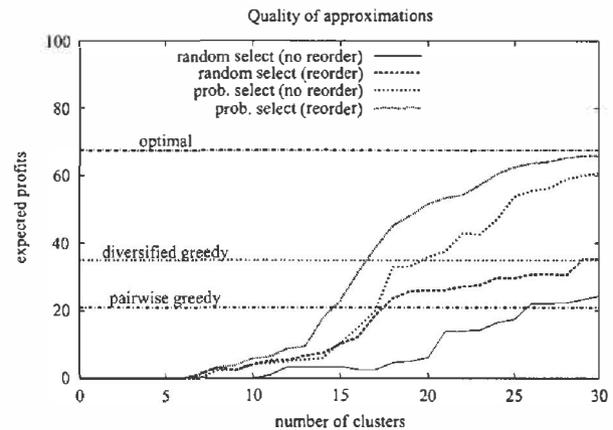

Figure 3: Average (lower) bound values (over 10 trials) and different seed selection and clustering methods. Horizontal lines show the optimal expected value and expected values for the diversified and pairwise greedy methods.

Although cluster-based approximations allow us to gradually improve the bound, ultimately, we are interested in finding the optimal assignment of $\mathbf{n}, \mathbf{m}$. Note that in such a case the optimal allocation may be obtained well before the value of a cluster-based approximation reaches the optimal value. Evaluating our experimental results in terms of allocations, we were able to find the optimal allocation in all 10 trials (considering up to 30 clusters) with the combination of two heuristics. Average number of clusters used to reach the optimal allocation was 22. Other methods missed the optimal allocations at least once. Random selection method with no reorder missed it in all trials.

## 6 Conclusions

Solving stochastic programming problems related to contract matching optimally requires to evaluate explicitly every possible combination of random variable values. To eliminate this dependency we focused on efficient heuristic approximations, in particular, a new clustering approach. Our primary contributions in this work include: a seed set clustering approach leading to upper and lower bound value estimates, and heuristics for finding good cluster-based approximations. The ability of our approach to solve succesfully hard contract matching problems was illustrated experimentally.

A number of new challenging research issues and questions emerge with our problem and need to be investigated; so-lutions or insights to some of them may further improve our current solutions. For example, at present our heuristics looks only at estimates of values and does not take any advantage of allocations obtained through upper and lower bound clusterings. The interesting question in this respect is whether there is any theory allowing us to detect portions of the optimal solution by examining upper and lower bound allocations, and whether there is a way to reduce the complexity of a problem by removing partial allocations known to be optimal.

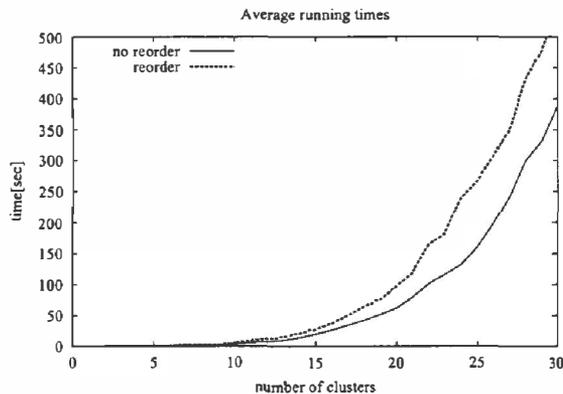

Figure 4: Average running times of approximations for different number of clusters.